\documentclass[final]{eptcs}

\usepackage[utf8]{inputenc}
\usepackage[T1]{fontenc}
\usepackage{amsmath,latexsym}
\usepackage{amsthm,amssymb}
\newtheorem{theorem}{Theorem}

\newtheorem{definition}[theorem]{Definition}
\newtheorem{example}[theorem]{Example}

\DeclareMathAlphabet{\mathitbf}{OML}{cmm}{b}{it}   

\newcommand{\titlerunning}{Counterfactual Explanations as Plans}
\newcommand{\authorrunning}{Vaishak Belle}

\title{Counterfactual Explanations as Plans}

\author{Vaishak Belle 
\institute{University of Edinburgh \& Alan Turing Institute, UK}
\email{vaishak@ed.ac.uk}}

\newcommand{\noact}{\langle\rangle}
\newcommand{\sub}{_}

\def\su{^}           

\newcommand{\theory}{\Sigma}
\newcommand{\theorydyn}{\theory\sub {\it dyn}}
\newcommand{\theoryinit}{\theory\sub 0}
\newcommand{\theoryinits}{\theory\sub 0 '}
\newcommand{\init}{\theoryinit}
\newcommand{\inits}{\theoryinits}
\newcommand{\dyn}{\theorydyn}

\newcommand{\D}{\theory}

\renewcommand{\P}{{\cal P}}

\newcommand{\R}{{\cal R}}

\newcommand{\W}{{\cal W}}

\newcommand{\Z}{{\cal Z}}

\newcommand{\abs}[1]{\left| #1\right|}
\newcommand{\set}[1]{\left\{ #1 \right\}}

\newcommand{\lan}{\langle}

\newcommand{\ran}{\rangle}

\newcommand{\poss}{{\it Poss}}

\newcommand{\oknow}{\mathitbf{O}}
\newcommand{\Bel}{\mathitbf{B}}
\newcommand{\know}{\mathitbf{K}}

\newcommand{\knows}{Knows}

\newcommand{\es}{{\mathcal E}{\mathcal S}}

\begin{document}
	
\maketitle

\begin{abstract} There has been considerable recent interest in explainability in AI, especially with black-box machine learning models.  As correctly observed by the planning community, when the application at hand is not a single-shot decision or prediction, but a sequence of actions that depend on observations, a richer notion of explanations are desirable. 

In this paper, we look to provide a formal account of ``counterfactual explanations," based in terms of action sequences. We then show that this naturally leads to an account of model reconciliation, which might take the form of the user correcting the agent's model, or suggesting actions to the agent's plan. For this, we will need to articulate what is true versus what is known, and we appeal to a modal fragment of the situation calculus to formalise these intuitions. We consider various settings: the agent knowing partial truths, weakened truths and having false beliefs, and show that our definitions easily generalize to these different settings. 

\end{abstract}

	\maketitle

\section{Introduction}

There has been considerable recent interest in explainability in AI, especially with black-box machine learning models, given applications in  credit-risk analysis,  insurance pricing and self-driving cars.  Much of this focus is on single-shot decision or prediction, but as correctly observed by the automated planning community \cite{fox2017explainable,cashmore2019towards,borgo2018towards}, in applications involving sequence of actions and observations and conditional plans that depend on observations, a richer notion of explanations are desirable. Fox et al.~\cite{fox2017explainable}, for instance, argue that understanding why a certain action was chosen (and not some other), why one sequence is more optimal than another, etc, are all desired constructs in explanations. 

Despite all this attention, there is yet to emerge a theory on how exactly to frame explanations in a general way. One candidate is the notable line of work on model reconciliation \cite{sreedharan2018handling}. The idea is that the agent might have incomplete or even false information about the world, and the user can advise the agent by correcting the agent’s model, or suggesting actions to the agent’s plan. The latter move is in the spirit of human-aware AI \cite{kambhampati2020challenges}. But such an idea is inherently \emph{epistemic}, and this brings the explainable AI literature closer to  \emph{epistemic planning} \cite{baral_et_al:DR:2017:8285}. Many  recent threads of work have tried to explicate this connection. 

In \cite{shvo2022resolving}, for example, the idea of discrepancy and resolving such discrepancy is studied. Using the epistemic logic fragment from \cite{muise-aaai-15}, where so-called proper epistemic knowledge bases are chosen for initial plan states that allow only for modal literals, discrepancy emerges when (in a dynamic logic-like language) \( \theory \models [\delta] \know\sub i \phi \) but \( \theory \not \models [\delta] \know\sub j \phi \). Given a goal \( \phi, \) an action sequence \( \delta, \) background knowledge \( \theory \) for agents \( i \) and \( j \) which include dynamic axioms, it turns out that \( i \) believes that \( \phi \) is true but \( j \) does not. So the resolution is to find some course of action \( \delta' \) that ensures that either \( \theory \models [\delta'] (\know \sub i \phi \land \know \sub j \phi) \) or that \( \theory \models [\delta'] (\know \sub i \neg \phi \land \know\sub j \neg \phi) \). That is, they both jointly believe after \( \delta' \) that \( \phi \) is made true, or that \( \phi \) is made false. And as pointed out in \cite{shvo2022resolving}, articulating the difference between beliefs and ground truth is needed for clarity. 

Likewise, the works on \emph{contrastive explanations} \cite{krarup2019model} as well as the credulous/skeptical semantics for model reconciliation from  \cite{vasileiou2022logic} are related. In the former, the ``why'' question is tackled by offering to add or remove actions from the current plan. In the latter, a notion of credulous and skeptical entailment is suggested as a way to keep updated mental states consistent in service of explanations.
Credulous entailment is when a single belief state suffices to check the validity and achievement of a plan, and skeptical entailment  when every belief state is involved.  The problem of ``explanation generation'', then,  between the user's theory \( \Sigma \) and the agent's theory \( \Sigma' \) is triggered when \( \Sigma\models [\delta] \phi \) for some goal \( \phi \) after sequence \( \delta \) but \( \Sigma' \not\models [\delta] \phi. \) This necessitates the updating of \( \Sigma' \) to \( \Sigma'' \)  such that \( \Sigma'' \models [\delta] \phi. \) Note that, even though  skeptical/credulous entailment involves belief states, the formalism itself is not epistemic. 
Perhaps it does not need to be in some limited cases, but ultimately a distinction between truth in the real-world and beliefs provides clarity on how the agent's model needs to be adjusted when plans do not work, either because its missing actions or entertaining incomplete/false truths. 

The reader may also surmise that there is clearly some relationship between these proposals, but it is not spelt out. In fact, no clear logical justification is given as why the definitions are reasonable in the first place.
As we mentioned, a theory on how exactly to frame explanations in such epistemic settings is yet to emerge.

%

What we seek to do in this paper is to develop and formalize ``counterfactual explanations'' over plans in the presence of both physical and sensing actions. Counterfactual explanations in machine learning are widely popular \cite{wachter2017counterfactual}, because they provide an intuitive account of ``what if'' and ``what could have been'', which helps us realize alternative worlds where a desired outcome might be achieved.\footnote{These notions will vary slightly in the dynamic version. We will focus on identifying the course of actions that can toggle the outcome (i.e., ``which plan"), and the information that is necessary to enable a  property (i.e., ``what should be known").}  Our account  can be seen as attempting to establish a logical relationship between contrastive explanations and knowledge based updates for model reconciliation and/or discrepancy. In fact, this relationship is a very simple one: it can be seen as a counterfactual! The ``why,'' ``what if'', and other such ``wh''-questions can be interpreted as counterfactuals \cite{pearl2009causality}. In the static setting, we are simply interested in an alterative world where some properties are different. In the dynamic setting, we might also be interested in an alternative plan that achieves a different outcome, very much in the spirit of contrastive plans. To our knowledge, there is no general account in the literature on counterfactual explanations as plans.

This brings us to the choice of the representation language. As already discussed above, we need to identify, in the first instance, a knowledge representation language for articulating things like what is true, what is known, whether something is not known, and whether something is falsely believed. This will give us an opportunity to formalize explanations in an epistemically adequate representation language. Although proposals such as \cite{muise-aaai-15,shvo2022resolving} might be perfectly reasonable, and might even be preferred for an automated planning account, but they do come with various syntactic stipulations that might affect the properties of the logic (e.g., disjunctive entailments). We believe the characterization is more easily stated and readability  improved by considering a general knowledge representation language, but nothing in the formalisation necessitates one choice of language over another. In fact, the underlying implementation can involve any of the recent proposals from the epistemic planning literature, e.g.,  \cite{muise-aaai-15,son2001formalizing}, of which we think \cite{shvo2022resolving} is particularly fitting. Our notion of counterfactual plans could be easily adapted from the algorithm of that work.

\textbf{Dimensions to formalisation} 
When attempting to formalise counterfactual (CF) explanations, as we shall show, there are multiple dimensions under which a definition can be explored. In the simplest case, given a plan \( \delta \) that achieves \( \phi \), a CF explanation might be a plan \( \delta' \) that negates \( \phi \). An interpretation could be as follows: if a plan ends up denying the loan to individual \( b, \) find  a plan that approves the loan for \( b \), but ensure that it is ``close'' to the original plan. This roughly captures counterfactuals in ML: given a data point \( x \) that has label \( y \), find  a data point \( x' \) that is minimally distant from \( x \) and has label  \( \neg y. \) We can further concretize this by explicating what closeness means, and whether additional  constraints can be provided so that a diverse range of \( x' \) are found \cite{mothilal2020explaining}. 

But when viewed through the lens of an interaction between an automated agent and a human user, in the sense of human-guided planning \cite{kambhampati2020challenges,sreedharan2018handling}, the framework becomes  richer. We will consider the case where the user assists in achieving goals. So this leads to a broader notion of counterfactuals: consider an alternate history or additional knowledge such that the goal now becomes true. This will require us to articulate the difference between what is true in the real world (the user's knowledge) versus what is believed (by the agent). Interestingly, this does not require a modelling language with multiple agents because the user's knowledge can serve  as proxy for truth in the real world.\footnote{This leads to a single-agent version formulation of, for example, the discrepancy condition from \cite{shvo2022resolving} in that it emerges whenever the user/root agent/real world is one where \( \theory \models [\delta] \phi \) but as far as the agent is concerned: \( \theory \not \models [\delta] \know \phi \). An account with multi-agent modal operators is possible in a straightforward way, using e.g., \cite{shapiro2002cognitive,DBLP:conf/kr/KellyP08} and \cite{Belle:2014aa} in particular, which is a many-agent extension to the language used in this paper.}
We consider various settings: the agent knowing partial truths, weakened truths and having false beliefs, and show that our definitions easily generalize to these different settings. The discrepancy model from \cite{shvo2022resolving}, the skeptical/credulous entailment model from \cite{vasileiou2022logic}, among others, can be seen as variations of this more general recipe in epistemic logic. We focus on the mathematical aspects here, but as mentioned, implementations from works such as \cite{shvo2022resolving} can be adapted for generating CF explanations as plans.


This then brings us to the choice of the formal language for our exposition. We choose the situation calculus \cite{reiter2001knowledge}, which has been well-explored for formalizing the semantics of planning problems \cite{pednault1989adl,reiter2001knowledge,Fritz:2008uq,DBLP:conf/aaai/Levesque96}. But because we are interested in nested beliefs, we explore a (newer) modal variant, the logic \( \es \) \cite{LakemeyerLevesque2004}, which provides a simpler semantics for planners \cite{classen2007towards}, as well closely mirrors the semantics of proposals such as  dynamic logic. Perhaps what makes it most interesting that under some conditions reasoning about actions and knowledge can reduce to non-modal (first-order or propositional) reasoning \cite{LakemeyerLevesque2004}, a feature we feel has not been considered extensively in the planning community. Our ideas 
do not hinge on this language, and so any planning language that helps us reason about truth, knowledge, actions and sensing should suffice.

\section{A Logic for Knowledge and Action} 
	\label{sec:reconstructing_the_epistemic_situation_calculus}

%

We now introduce the logic \( \es \) \cite{LakemeyerLevesque2004}.\footnote{Our choice of language may seem unusual, but it is worth noting that this language is a modal syntactic variant of the classical epistemic situation that is better geared for reasoning about knowledge \cite{Lakemeyer:2011:SCU:1897346.1897552}. But more importantly, it can be shown that reasoning about actions and knowledge reduces to first-order reasoning via the so-called regression and representation theorems \cite{LakemeyerLevesque2004}. (For space reasons, we do not discuss such matters further here.) There are, of course, many works explicating the links between the situation calculus and logic programming; see \cite{reiter2001knowledge} for starters.} It is an epistemic logic, but we only need the objective fragment for formalizing counterfactual explanations as plans. When we consider the more elaborate notion of reconciliation-type explanations where both a user and an agent is necessary, we will use the full language.

	The non-modal fragment of \( \es \)  consists of standard first-order logic with $=$. That is, connectives \( \set{\land, \forall, \neg} \),  syntactic 
	  abbreviations \( \set{\exists,\equiv, \supset} \) defined from those connectives, and a supply of variables variables \( \set{x,y,\ldots, u, v, \ldots} \). Different to the standard syntax, however, is the inclusion of (countably many) \emph{standard names} (or simply, names) for both objects and actions \(\R,  \)  which will allow a simple, substitutional interpretation for \( \forall \) and \( \exists. \) These can be thought of as special extra constants that satisfy the unique name assumption and an infinitary version of domain closure.
  
	Like in the situation calculus, to model immutable properties, we assume rigid predicates and functions, such as {\it $IsPlant(x)$} and {\it $father(x)$} respectively. To model changing properties, \( \es \) includes fluent predicates and functions of every arity, such as {\it $Broken(x)$}  and {\it $height(x)$}. Note that there is no longer a situation term as an argument in these symbols to distinguish the fluents from the rigids. For example, \( \es \) also includes a distinguished  fluent predicates \( \poss \) and \( SF \) to model the executability of actions and capture sensing outcomes respectively, but they are now a unary predicates. Terms and formulas are constructed as usual. The set of ground atoms \( \P \) are  obtained by applying all object names in \( \R \) to the predicates in the language. 

	There are four modal operators in \( \es \): \( [a], \Box, \know \) and \( \oknow. \) For any formula \( \alpha, \) we read \( [a]\alpha, \Box\alpha \) and \( \know\alpha \) as ``$\alpha$ holds after $a$", ``$\alpha$ holds after any sequence of actions" and ``\( \alpha \) is known,'' respectively. Moreover, \( \oknow\alpha \) is to be read as ``\( \alpha \) is only-known.'' Given a sequence \( \delta = a\sub 1 \cdots a\sub k, \) we write \( 
[\delta]\alpha \) to mean \( [a\sub 1] \cdots [a\sub k]\alpha. \) We write \( a\cdot \delta \cdot a' \) to mean \( [a]\cdot [a\sub 1] \cdots [a\sub k] \cdot [a']. \)
	
	In classical  situation calculus parlance, we would use \( [a]\alpha \) to capture successor situations as properties that are true after an action in terms of  the current state of affairs. 
Together with the \( \Box \) modality, which  allows to capture quantification over situations and histories, basic action theories can be defined. Like in the classical approach, one is interested in the entailments of the basic action theory. \\

	\textbf{Semantics} 
	\label{sub:semantics}
	Recall that in the simplest setup of the possible-worlds semantics, worlds mapped propositions to \( \set{0,1} \), capturing the (current) state of affairs. \( \es \) is based on the very same idea, but extended to dynamical systems. So, suppose a world maps \( \P \) and \( \Z \) to \( \set{0,1} \).\footnote{We need to extend the mapping to additionally interpret fluent functions and rigid symbols, omitted here for simplicity.} Here,  \( \Z \) is the set of all finite sequences of action names, including the empty sequence \( \lan\ran. \) Let \( \W \) be the set of all worlds, and \( e\subseteq \W \) be the \emph{epistemic state}. By a \emph{model}, we mean a triple \( (e,w,z) \) 
	 where \( z\in \Z. \)

	Intuitively, each world can be thought of a situation calculus tree, denoting the properties true initially but also after every sequence of actions. \( \W \) is then the set of all such trees. 
	Given a triple \( (e,w,z) \), \( w \) denotes the real world, and \( z \) the actions executed so far. Interestingly, 
	\( e \) captures the accessibility relation between worlds, but by modeling the relation as a set, we are enabling positive and negative introspection using a simple technical device. 



	To account for how knowledge changes after (noise-free)  sensing, one defines \( w' \sim\sub z w \), which is to be read as saying  ``\( w' \) and \( w \) agree on the sensing for \( z \)'', as follows:

	\begin{itemize}
		\item if \( z=\lan\ran, \) \( w' \sim\sub {z} w \) for every \( w' \); and
		 \item \( w'\sim \sub {z\cdot a} w \) iff \( w'\sim\sub {z} w, \) and \( w'[SF(a),z] = w[SF(a),z]. \)
	\end{itemize}
	

	This is saying that initially, we would consider all worlds compatible, but after actions, we would need the world \( w' \) to agree on sensing outcomes. The reader might notice that this is clearly a reworking of the successor state axiom for the knowledge fluent in \cite{citeulike:528170}.

	With this, we get a simply account for truth. We define the satisfaction of formulas wrt the triple \( (e,w,z) \), and the semantics is defined inductively:
	 \begin{itemize}
		\item \( e,w,z\models p \) iff \( p \) is an atom and \( w[p,z] =1 \);
		\item \( e,w,z \models \alpha\land\beta \) iff \( e,w,z\models \alpha \) and \( e,w,z\models \beta; \) 

		\item \( e,w,z\models \neg\alpha \) iff \( e,w,z\not\models \alpha; \) 

		\item \( e,w,z\models \forall x\alpha \) iff \( e,w,z\models \alpha^x_n \) for all \( n \in \R; \)

		\item \( e,w,z\models [a]\alpha \) iff \( e,w,z\cdot a\models \alpha; \)
			\item \( e,w,z\models \Box \alpha \) iff \( e,w,z\cdot z' \models \alpha \) for all \( z' \in \Z \); 
		
		\item \( e,w, z\models\know\alpha \) iff for all \( w' \sim \sub z w, \) if \( w'\in e, \) \( e,w',z\models\alpha \);  
		
		 \item \( e,w, z\models\oknow\alpha \) iff for all \( w' \sim \sub z w, \)  \( w'\in e, \) iff \( e,w',z\models\alpha \). 
	
	\end{itemize} 
  
	To define entailment for a logical theory, we write \( \D \models \alpha \) (read as ``$\D$ entails $\alpha$'') to mean for every \( M= (e,w, \lan\ran) \), if \( M\models \alpha' \) for all \( \alpha' \in \D,  \) then \( M\models \alpha. \) We write \( \models \alpha \) (read as ``$\alpha$ is valid'') to mean \( \set{} \models \alpha. \) \\

	\textbf{Properties} 
	\label{sub:properties} 
	Let us first begin by observing that given a model \( (e,w,z), \)  we do not require \( w\in e.  \) It is easy to show that if we stipulated the inclusion of the real world in the epistemic state, \( \know\alpha\supset \alpha \) would be true. That is, suppose \( \know \alpha. \) By the definition above, \( w \) is surely compatible with itself after any \( z \), and so \( \alpha \) must hold at \( w. \) Analogously, properties regarding knowledge can be proven with comparatively simpler arguments in a modal framework, in relation to the classical epistemic situation calculus. Valid properties include: (a) \(  \Box(\know(\alpha)\land \know(\alpha\supset \beta)\supset \know(\beta)) \);  (b) \(  \Box(\know(\alpha) \supset \know(\know(\alpha))); \) (c) $\Box(\neg\know(\alpha) \supset \know(\neg \know(\alpha)))$; (d) $\Box(\forall x.~\know(\alpha) \supset \know(\forall x.~\alpha))$; and
	(e) $\Box(\exists x.~\know(\alpha) \supset \know(\exists x.~\alpha)).$

	Note that such properties hold over all possible action sequences, which explains the presence of the \( \Box \) operator on the outside. The first is about the closure of modus ponens within the epistemic modality. The second and third are on positive and negative introspection. The last two reason about quantification outside the epistemic modality, and what that means in terms of the agent's knowledge. For example, item 5 says that if there is some individual \( n \) such that the agent knows \( Teacher(n) \), it follows that the agent believes \( \exists x Teacher(x) \) to be true. This may seem obvious, but note that the property is really saying that the existence of an individual in some possible world implies that such an individual exists in all accessible worlds. It is because there is a fixed domain of discourse that these properties come out true; they are referred to a the  Barcan formula. 
	
	It is worth nothing that in single-agent epistemic planning \cite{baral_et_al:DR:2017:8285}, it is most common to have epistemic goals of the sort \( \know\phi \), \( \neg \know \phi \) and \( \know\neg\know\phi \), where \( \phi \) is non-modal. The idea is that we might be interested in interleaving physical and sensing actions such that (respectively) \( \phi \) becomes known, or as an observer (e.g., user) we make note that the agent does not know \( \phi \), or that the agent knows that it does not know \( \phi \), in which case it might choose to do actions so that it gets to know \( \phi \). Multiple nestings of modalities are allowed but usually not necessary in the single-agent case. When multiple agents are involved, however, \cite{DBLP:conf/kr/KellyP08,muise-aaai-15,Belle:2014aa}, it becomes necessary to interleave epistemic operators, often arbitrarily, in service of notions such as common knowledge \cite{DBLP:journals/jacm/HalpernM90}. 
	
	 As seen above, the logic \( \es \) allows for a simple definition of the notion of only-knowing in the presence of actions \cite{77758},  which allows one to capture both the beliefs as well as the non-beliefs of the agent. Using the modal operator \( \oknow \) for only-knowing, it can be shown that \( \oknow\alpha\models \know\beta \) if \(\alpha\models\beta \) but \( \oknow\alpha\models \neg \know\beta \) if \( \alpha\not\models\beta \) for any non-modal \( \set{\alpha,\beta}. \) That is, only-knowing a knowledge base also means knowing everything entailed by that knowledge base. Conversely, it also means not believing everything that is not entailed by the knowledge base. In that sense, \( \know \) can be seen as an ``at least'' epistemic operator, and \( \oknow \) captures both at least and ``at most" knowing. This can be powerful to ensure, for example, that the agent provably does not know protected attributes. 

	We will now consider the axiomatization of a basic action theory in \( \es \). But before explaining how successor state axioms are written, one might wonder whether a successor state axiom for \( \know \) is needed, as one would for \( \knows \) in the epistemic situation calculus. It turns out because the compatibility of the worlds already accounted for the executability of actions and sensing outcomes in accessible worlds, such an axiom is actually a property of the logic:  \begin{align*}\label{eq:kssa}
		\models \Box[a]\know(\alpha) \equiv 
			(SF(a) \land \know(SF(a) \supset [a]\alpha)) ~\lor    (\neg SF(a) \land \know(\neg SF(a) \supset [a]\alpha)).
	\end{align*} 
%

({Free variables are implicitly quantified from the outside.}) 
What will be known after an action is 
based on what is true in the real world and the incorporation of this information with the agent's knowledge. \\

	
	




\textbf{Basic Action Theories} 
To illustrate the language towards the axiomatization of the domain, we consider the analogue of the basic action theory in the situation calculus \cite{reiter2001knowledge}. It consists of: \begin{itemize}
	\item axioms that describe what is true in the initial states, as well as what is known initially; 
	\item  precondition  axioms that describe the conditions under which actions are executable using a distinguished predicate \( Poss \);  
	\item  successor state axioms that describe the conditions under which changes happen to fluents on executing actions,  incorporating Reiter's monotonic solution to the frame problem; and 
	\item sensing axioms that inform the agent about the world using a distinguished predicate \( SF. \)
\end{itemize}

Note that foundational axioms as usually considered in Reiter's variant of the situation calculus \cite{reiter2001knowledge} are not needed as the tree-like nature of the situations is baked into the semantics.

We will lump the successor state, precondition and sensing axioms as \( \theorydyn \). The sentences that are true initially will be referred to by \( \theoryinit \). When we are not interested in epistemic goals, and do not concern ourselves with sensing actions, we can restrict our attention to entailments of \( \init \land \dyn. \) Note that because \( \init \) might include disjunctions (and possibly quantifiers), there might be multiple worlds where \( \init \) is true. For example, in a propositional language with only two propositions \( \set{p,q} \), \( \init = (p\land \neg q) \)  means there is only a single world where \( \init \) is true initially, but \( \init = (p \lor q) \)  means that there are three worlds where \( \init \) is true initially. In other words, \( \init \) might correspond to a single or multiple initial states in classical planning parlance. 

 %

If we are wanting to model knowledge, the agent cannot be expected to know everything that is true, and so let \( \theoryinits \) be what is believed initially. It may seem natural to let \( \theoryinits \subseteq \theoryinit \), but that it not necessary. The agent might  be uncertain about what is true (e.g., \( \theoryinit \) might have \( p \) but \( \theoryinits  \) has \( p\lor q \) instead).\footnote{If the agent believes facts that are conflicted by observations about the real world, beliefs may need to be revised \cite{Delgrande:2012fk}, a matter we ignore for now. Our theory of knowledge is based on \emph{knowledge expansion} where sensing ensures that the agent is more certain about the world \cite{citeulike:528170,reiter2001knowledge}. In the case of reconciliation-based explanations, however, we will need to entertain a simple type of revision based on the deletion of facts from \( \theoryinits, \) as we shall shortly see. A general treatment of deleting in first-order languages might be based on \emph{forgetting}  \cite{lin1994forget}.} However, for simplicity, we will require that agents at least believe the dynamics works as would the real world. Therefore, we consider entailments wrt the following \emph{background theory}: \begin{equation}\label{eq:example}
	\theory = \theoryinit \land \theorydyn \land \oknow(\theoryinits \land \theorydyn).
\end{equation} 

There are conveniences afforded by a basic action theory of this form. Firstly, as far as a non-epistemic account of planning is concerned (that is, one where the knowing modality is not present in the goal), we would be  checking the entailment of non-modal goal formulas, and therefore, it is immediate that only \( \theoryinit \land \theorydyn \) from \( \theory \) is involved. Everything in the context of an epistemic operator in \( \theory \) can be ignored. This is precisely what we will explore in the first set of results on counterfactual plans. But when we need to refer to formulas involving epistemic modalities, we will only need to refer to the non-objective parts of \( \theory. \) (When the agent performs sensing actions, however, they will provide values from \( \theoryinit. \)) Thus, we can simply concern ourselves with entailments of \( \theory \) henceforth. \\


%
%

\textbf{Example} 
\label{sub:example}
	Let us consider a simple blocks world example, involving picking up and dropping objects, but also quenching (rapidly cooling to very low temperatures) objects so that they become fragile, adapted from \cite{DBLP:journals/sLogica/LesperanceLLS00,LakemeyerLevesque2004}. As usual, picking up is only when possible when the robot is not already holding anything, and dropping is only possible when it is already holding the object. Also, broken objects in the robot's hand can be repaired. So, \begin{align*}
		 \Box Poss(a)  \equiv~ & (a=pickup(x) \land \forall z.\neg Holding(z)) \lor  (a=drop(x)\land Holding(x)) \lor \\ & (a=quench(x)\land Holding(x)) \lor  (a=repair(x) \land Holding(x) \land Broken(x)). 
	\end{align*} 
Let us also permit a sensing axiom that allows one to look up if an object is made of glass: \[
	\Box SF(a) \equiv (a=isGlass(x) \land Glass(x)) \lor a \neq isGlass(x).
\]

To now consider successor state axioms, let us suppose holding an object is possible by picking it up. A fragile object gets broken on dropping it, and not repairing it. Quenching makes an object fragile, regardless of whether it was previously fragile or not. An object being a glass is a rigid property. These are formalized as the axioms below, where the left hand side of the equivalence captures the idea that for every sequence of actions, the effect of doing \( a \) on a predicate is given by the right hand side of the equivalence.   These capture Reiter's monotonic solution to the frame problem using successor state axioms, but now in \( \es. \)
\begin{align*}
	 \Box [a]& Holding(x) \equiv   a=pickup(x) \lor     (Holding(x) \land a \neq drop(x)).  \\
	\Box [a] & Broken(x) \equiv  (a=drop(x) \land Fragile(x)) \lor    (Broken(x) \land a \neq repair(x)).		\\ 
	\Box [a] & Fragile(x) \equiv   Fragile(x) \lor a =quench(x). \\
		\Box [a]   & Glass(x) \equiv Glass(x).
\end{align*}

Let us suppose the initial theory only-believed by the agent is the following, where nothing is held, there is a non-broken object \( c \) made of glass, 
and as one would assume, glass objects are fragile: \begin{align*}
	\inits = & \{ Glass(c), \neg \exists x Holding(x), \neg Broken(c),  \forall x (Glass(x) \supset Fragile(x)) \}. 
\end{align*}
In the real world, let us additionally suppose there is another glass object \( d \) but also a non-fragile object \( h \):\footnote{We use the set notation and the formula notation (that is, using conjunctions of formulas) for  theories as per convenience.} \(
	\init = \inits \cup \set{(Glass(d)),  (\neg Glass(h)), (\neg Fragile(h))}.
\) 

That is, whatever the agent believes happens to be true in the real world, but the agent does not know about \( d \) being made of glass and \( h \) not being fragile. \( \init \) in itself does not commit to how many objects there are in the universe, and so it should be clear to the agent that there are (possibly infinitely) many  objects outside of \( c \) for which it is not known whether they are fragile or made of glass, for example. 

Here a few examples of entailments of \( \theory \): (a) \(  (\neg \know Glass(d) \land \neg \know \neg Glass(d)) \); (b) \(   \know \neg \know Glass(d) \); (c) \(   [isGlass(h)] \know  Glass(d) \); and (d) \(  [isGlass(h)] \know \know \know Glass(d). \)

That is, the agent’s initial beliefs imply that the agent does not know whether \( d \) is made of glass. Moreover, by introspection, the agent knows that it does know if 	\( d \) is made of glass. But after sensing $h$ for glass, it knows that it knows that it knows (and so on arbitrarily) that $d$ is made of glass.

\section{Reasoning \& Planning} 
\label{sub:reasoning}

Given a background theory \( \theory \), an action sequencen \( \delta = a\sub 1 \cdots a\sub k, \) and a (non-modal) goal formula \( \phi \), the classical problem of \emph{projection} \cite{reiter2001knowledge} is to identify if the sequence enables \( 
\phi \). That is, whether
\(
	\theory \models [\delta]\phi.
\) 
We also want to ensure that the action sequence is executable (aka \emph{valid} and/or \emph{legal}). So let us \( Exec(\noact) = true \), and \( Exec(a\cdot \delta) = Poss(a) \land [a]Exec(\delta) \). Then, we check: \(
	\theory \models Exec(\delta) \land  [\delta]\phi.
\) 
In the epistemic setting \cite{citeulike:528170}, we are interested in checking if \( \phi \) is known after executing \( \delta \), which might include sensing actions too. That is, whether\footnote{It might also be of interest to know whether \( \phi \) is true \cite{fan2015contingency}, that is, checking that \(
	\theory \models [\delta] (\know \phi \lor \know \neg \phi).
\) 
Here, the second disjunct is asserting that the agents knows \( \phi \) to be false.} \(
	\theory \models [\delta]\know \phi. 
\)
When adding action executability, we would have: \(
	\theory \models [\delta]\know\phi \land \know Exec(\delta).
\) 
So the agent also knows that the sequence is executable: which means \( \delta \) is executable in every world in the agent's epistemic state.
Note that because we do not require the real world to be necessarily included  in the epistemic state, it is not necessarily that \( \delta \) is actually executable in the real world. If we needed to additionally enforce that, we would need: \(
	\theory \models Exec(\delta) \land [\delta]\know\phi \land \know Exec(\delta).
\)

The task of planning, then, is to identify a sequence \( \delta \) such that \( \phi \) is made true (in the non-epistemic setting) or that \( \phi \) is known to be true, and that the appropriate executability condition holds. 



Note that, we do not require plans to simply be a sequence of (physical) actions. For one thing, they may involve sensing actions, based on which the agent obtains information about the world. For another, we might be interested plans involving recursion \cite{siddthesis}, conditional statements and tests \cite{DBLP:conf/aaai/Levesque96,Levesque97-Golog,Fritz:2008uq}. Such plan structures do not change the nature of the reasoning problem, however: no matter the plan structure, we will be evaluating if the sequence of actions executed by the agent enables some goal, that is, whether the structure instantiates a sequence \( \delta \) such that \( \theory \models [\delta]\phi \) and \( \theory\models [\delta] \know \phi \) for world-state and epistemic planning respectively. Likewise, regression is not limited to only action sequences and can work with conditional and recursive plans.\footnote{In the expressive programming formalism of {\sc GOLOG} \cite{Levesque97-Golog}, for example, we provide the semantics for program execution such that there is a history (an action sequence) that terminates the program in addition to satisfying the goal \cite{Fritz:2008uq}. Likewise, with loopy plans, we provide the semantics for plan execution such that there is a history that reaches the final state of the plan structure in addition to  goal satisfaction \cite{DBLP:conf/aaai/Levesque96}.}

Finally, it can be shown that reasoning about actions and knowledge can be reduced to non-modal reasoning.  We omit the details but refer readers to \cite{LakemeyerLevesque2004}. (We will included an extended report with some examples.) With a finite domain assumption, this can be further reduced to propositional reasoning. 

%
%
%
%
%


\section{Counterfactual Explanations} 
\label{sec:counterfactual_e}

We will firstly attempt to characterize counterfactual (CF) explanations as plans, at an objective level. This could be viewed, therefore,  as an instance of planning with incomplete information,  but it can also be ultimately linked to the epistemic setting, as we shall see below. Simply put,  a CF explanation is an alternative course of action that negates the goal.  (Conversely, if some sequence does not enable the goal, we search for an explanation that does.\footnote{In relation to machine learning \cite{wachter2017counterfactual}, the idea is to produce an action sequence that changes the outcome. For example, if \( \phi \) represents an applicant getting rejected for a job application, then we find a plan to ensure that they are accepted. Conversely, if \( \phi \) states that moving an object to a different room causes it to break, we find a plan to ensure that the object is not broken during the move. In another scenario, if \( \phi \) states that high-risk individuals have their loan approved, we might be interested in additional assumptions that ensure that such individuals are mostly denied loans unless further constraints hold. 

Thus, it is not the polarity of the formula that is relevant here, and our use of the term “goal” is perhaps slightly misleading. Essentially, our definitions below formalize the identification of conditions and sequences that toggle the outcome.})


\begin{definition} Suppose \( \theory \models Exec(\delta) \land [\delta]\phi. \) 
	 A CF explanation for \( \phi \) after \( \delta \) is an action sequence \( \delta' \) such that \( \theory \models Exec(\delta') \land  [\delta'] \neg \phi \) and \( dist(\delta',\delta) \) is minimal.
	
\end{definition}

One natural candidate for the distance metric is the cost of actions (and hence the cost of plans) \cite{vasileiou2022logic}. Let us explore some  measures below that does not necessitate associating explicit numbers with actions for simplicity. Of course, the appropriate measure might very well depend on the application domain. 

\begin{definition} Given two sequences \( \delta, \delta' \), define length-based minimality as minimizing 
	
	\( \abs{(length(\delta') - length(\delta))} \), which is the absolute value of the difference in lengths. Length is defined inductively: \( length(\noact) = 0 \), and $length(\delta \cdot a) = length(\delta) + 1$. 
\end{definition}

\begin{example} Suppose \( \delta = pickup(c)\cdot drop(c) \cdot repair(c) \) and the goal is \(  \neg Broken(c) \). The shortest CF explanation for \( Broken(c) \) is \( \delta' = pickup(c)\cdot drop(c) \). That is: (a)
	\( \theory \models [\delta] \neg Broken(c) \); and  (b) \( \theory \models [\delta'] Broken (c). \) 
\end{example}

Let us consider another measure based on  all the properties of the world that are affected. 
For any \( \delta \), define \( fluents(\delta) \) as the set of all  fluents mentioned in the successor state and precondition axioms of actions in \( \delta. \)

\begin{definition} Given \( \delta, \delta' \) as above, define  fluent-based minimality as minimizing
	
	 \( \abs{ size(fluents(\delta)) - size(fluents(\delta'))} \). 	
\end{definition}

\begin{example} Suppose \( \delta = pickup(h)\cdot drop(h) \) for goal \( \neg Broken(h) \). The fluent set for \( \delta \) is 
	
	\( \set{Holding(x), Broken(x)} \). Because \( h \) is not fragile, we would need to quench it. Consider that the fluent set for \( \delta' = pickup(h)\cdot  quench(h)\cdot drop(h) \) is \( \set{Holding(x), Fragile(x), Broken(x)} \), and so it is minimally larger: that is, there is no other \( \delta' \) with the same fluent set as \( \delta \) which achieves \( Broken(h) \). As desired, \( \theory \models [\delta'] Broken(h) \).
	
\end{example}

As it turns out, only optimizing for the affected set is not quite right because many irrelevant ground actions could be included.

%

\begin{definition} Given \( \delta, \delta' \) as above, define plan-and-effect minimality as jointly minimizing both length-based and fluent-based minimality. 
	
\end{definition}

\begin{example} It should be clear that only optimizing for fluent-based minimality is problematic. Consider once more, \( \delta = pickup(h)\cdot drop(h) \) for goal \( \neg Broken(h) \). Let \( \delta'' = pickup(d)\cdot   drop(d) \cdot \delta' \),  where \( \delta' = pickup(h)\cdot  quench(h)\cdot drop(h). \) The fluent set of \( \delta'' \) does not differ from that of \( \delta \) much more than that of \( \delta' \) does. However, \( \delta'' \) has some irrelevant actions for achieving \( Broken(h) \). Thus, \( \delta' \) achieves plan-and-effect minimality.
	
\end{example}

An important additional ingredient with counterfactual explanations is \emph{diversity} \cite{mothilal2020explaining}, where we might seek multiple CF explanations but constrained according some feature. For example, we could be looking for students whose scored low in mathematics (the constraint) while still graduating (the latter being the goal), looking for tall students (the constraint) who still do not play basketball well (the goal), and so on. Properties such as people being tall can be modelled as rigid predicates, but our definition does not limit itself to rigids.


\begin{definition} Given \( \theory, \delta, \phi \) as above, \( k\in \mathbb N, \) and any non-modal formula \( \alpha \) as a diversity constraint, a diverse CF explanation is a sequence \( \delta' \) such that \( \theory \models Exec(\delta') \land [\delta'](\alpha \land \neg \phi) \)	and \( dist(\delta',\delta) \leq k. \) 
\end{definition}

\begin{example} Suppose \( \delta = pickup(h)\cdot drop(h) \) for goal \( \phi = \exists x \neg Broken (x) \). Suppose we are interested in a broken object, but with the diversity constraint of it being made of glass. In other words, \( \alpha = \exists x Glass(x) \), and so we are to find a sequence \( \delta' \)  such that \( \exists x (Glass(x)\land Broken(x)) \) is made true. It is easy to see that \( \delta' = pickup(c)\cdot drop(c) \) is such an explanation. 	
\end{example}

This then leads to multiple explanations that are close enough. 

\begin{definition} Let \( k \) be any positive integer denoting the closeness upper bound. Given \( \theory, \delta, \phi \) as above, and any non-modal formula \( \alpha \) as a diversity constraint, diverse CF explanations is a set of sequences \( \set{\delta\sub 1, \ldots, \delta\sub n} \) such that \( \theory \models Exec(\delta\sub i) \land  [\delta\sub i](\alpha \land \neg \phi) \) for every \( i \)	and \( dist(\delta\sub i,\delta) \leq k \). 
	
\end{definition}

\section{Reconciliation-based Explanations} 
\label{sec:reconciliation_based_explanations}

The simplest case of an agent providing a counterfactual explanation is that we formulate plans in the context of knowledge, and so goals can involve nested beliefs. 

\begin{definition} Suppose 
	\( \theory \models [\delta] \know \phi \land \know Exec(\delta) \), where \( \phi \) might mention \( \know \) but no other modality. Then a counterfactual explanation is \( \delta' \)  such that \( dist(\delta',\delta) \) is minimal and \( \theory \models [\delta'] \know \neg \phi \land \know Exec(\delta'). \) 
	
\end{definition}

Recall that we are not seeking \( [\delta] \neg \know \phi \), because this is  the case of an agent being ignorant. We instead seek  \( \delta' \) after which the agent knows that \( \phi \) is false. 

Note that in the above definition we were not stipulating executability in the real world, because it suffices for the account to be purely epistemic. We can enforce this additionally, of course, but it will come up naturally for the definitions below because the user needs to make sure she is only suggesting legal actions.

\begin{example} Following our examples above, consider \( \delta = pickup(c)\cdot drop(c)  \) and clearly \( \theory \) entails \( [\delta]\know Broken(c)\). The explanation \( \delta' = \delta\cdot repair(c) \) achieves \( \know\neg Broken(c). \)
	
\end{example}

What we will consider below is the case where the user assists in achieving goals. So this leads to a broader notion of counterfactuals: consider an alternate history or additional knowledge such that the goal becomes true.

\subsection{Agents Only-Knowing Partial Truths} 
\label{sub:ignorant_agents}

For this case of ignorant agents, we  assume \( \theoryinits \subseteq \theoryinit \).\footnote{As mentioned before, we do not require \( \Box(\know\alpha\supset\alpha) \) to be valid, but if this was stipulated in the logic (by insisting that the real world \( w\in e \)), then it should always be that \( \init\models\inits \). (If not, then \( \know\inits\supset \inits \) would be falsified in the real world.)}  Suppose a plan \( \delta \) fails in achieving \( \know\phi. \) A CF explanation here amounts to considering a possible world and a sequence such that the agent knows \( \phi \). So either there are missing actions, or missing knowledge, or both.
 

\begin{definition} (Missing actions.) Suppose \( \theory \models Exec(\delta) \land \know Exec(\delta) \) but  \( \theory \not \models [\delta] \know \phi \), that is, \( \theory \models [\delta] \neg \know \phi. \) Suppose there is a sequence \( \delta' \) such that \( dist(\delta',\delta) \) is minimal,  \( \theory \models [\delta'] \phi \land Exec(\delta'), \) and \( \theory \models [\delta'] \know \phi \land \know Exec(\delta'). \) Then a CF explanation is \( \delta' \).
\end{definition} 

Note that we do not insist \( \theory \models [\delta] \phi \), because as the definition title suggests, there might actions missing. 
One might also wonder why we insist on the agent also needing to know \( \phi \) after \( \delta' \): is it not redundant? The answer is no. Firstly, notice that even if there is \( \delta' \) such that \( \theory\models[\delta' ] \phi \), it is not necessary that \( \delta' \) is minimally away from \( \delta. \) For example, as far as entailment of objective formulas is concerned, the presence of sensing actions in \( \delta' \) is irrelevant: sensing only affects the knowledge of the agent and does not affect the real world. But the agent may very well need sensing actions to learn more about the world. Therefore, we insist that we need to find a \( \delta' \) that is minimally different to \( \delta, \)  enables \( \phi \) in the real world but also enables the knowing of \( \phi \). In other words, if both \( \delta' \) and \( \delta'' \) enable \( \phi \) and they differ only in that \( \delta' \) includes sensing actions whereas \( \delta'' \) does not, then we want \( \delta' \) to be the explanation. 

\begin{example} Consider \( \delta = pickup(c) \) for the goal \( \neg \know Broken(c) \). The explanation is \( \delta' = \delta \cdot drop(c) \), and indeed, \( \theory \models [\delta'] Broken(c) \) but also \( \theory \models [\delta'] \know Broken(c) \). 
	
\end{example}

\begin{example} Suppose \( \delta = pickup(d)\cdot drop(d) \) for the goal \( \neg \know Broken(d) \). But in fact, \( \theory \models [\delta] Broken(d) \), and so the agent does not know \( d \) is broken owing to the fact that it does not know that \( d \) is made of glass. So consider \( \delta' = pickup(d)\cdot isGlass(d) \cdot drop(d) \). This does not affect what is true in the world, but does lead the agent to know that \( Glass(d) \). Therefore, \( \delta' \) is the explanation since \( \theory \models [\delta'] Broken(d) \), and \( \theory\models [\delta'] \know Broken(d) \). 
	
\end{example}

\begin{definition} (Missing knowledge.) Suppose \( \theory \models Exec(\delta)  \). 
	 Suppose \( \theory \not\models [\delta] \know \phi \land \know Exec(\delta) \) but \( \theory \models [\delta]\phi. \) 
	Then suppose there is some \( \alpha \in \theoryinit - \theoryinits \) such that \( \init \land \dyn \land \oknow(\inits \land \alpha \land \dyn) \models [\delta]\know\phi \land \know Exec(\delta). \) Then the smallest such \( \alpha \) is the explanation. 
	
\end{definition}

Note that we do not assume in the definition that  \( \theory \models  \know Exec(\delta) \), because such an \( \alpha \) could be necessary knowledge to reason about the executability of actions.


\begin{example} Given \( \delta = pickup(d)\cdot drop(d) \), we know that \( [\delta] \neg \know Broken(d) \). 
	%
	But for \( \alpha = Glass(d) \), and owing to the fact that every object made of glass is declared to be fragile in \( \inits \), 
	we see that \( \init\land \dyn \land \oknow(\inits\land\alpha\land \dyn) \) entails \( [\delta] \know Broken(d) \). So \( \alpha \) is the explanation. 
	
\end{example}

\begin{definition}\label{defn missing knowledge and action} (Missing knowledge and action.) Suppose \( \theory \models Exec(\delta) \) but \( \theory \not\models [\delta] \phi, \)  or 
	\( \theory \not\models [\delta]\know\phi \). Suppose there is a minimally distant \( \delta' \) and some \( \alpha \) such that \( \init \land \dyn \land \oknow(\inits \land \alpha \land \dyn)  \) entails: \(  [\delta'](\phi \land \know\phi) \land Exec(\delta') \land \know Exec(\delta'). \) Then the smallest such \( \alpha \) together with \( \delta' \) is the explanation.
	
	
\end{definition}

Note that, firstly, if \( \Box (\know\alpha\supset \alpha) \) was true in the logic, \( \theory \not\models [\delta] \phi \) also means \( \theory \not\models [\delta] \know \phi \), because the real world \( w\in e \). By assumption, if \( \phi \) is not made true after \( \delta, \) then the agent cannot come to know \( \phi \) after \( \delta. \) Since we do not require knowledge being true, the definition has to make stipulations about both \( [\delta]\phi \) and \( [\delta]\know \phi. \) Moreover, it is possible that \( \theory\models [\delta] \phi \) but \( \theory\not\models [\delta]\know \phi \)  because there are  sensing actions that could enable the agent to learn sufficient information for knowing  \( \phi \)  after \( \delta \), or because there is some information that cannot be accessed by sensing that needs to be added to \( \inits \) for the agent to infer \( \know\phi \) after \( \delta, \) (or both).\footnote{This suggests a ``criteria'' for triggering the addition of knowledge. Define two sequences \( \delta \) and \( \delta' \) to be close iff \( \delta' \) only differs from \( \delta \) in having sensing actions. Condition knowlege addition only when there is no \( \delta' \) that is close in this sense, and \( \theory \models [\delta]\phi \land [\delta] \neg \know \phi \land [\delta'] \know \phi \) along with \( \theory \models Exec(\delta) \land Exec(\delta') \land \know Exec(\delta) \land \know Exec(\delta') \). So,  if there is a legal sequence that only augments \( \delta \) with sensing  but enables knowing the goal, then we conclude that no new knowledge needs to be added. If there are multiple such augmented sequences \( \delta'  \) and \( \delta'' \), we would choose the shortest such sequence. This would avoid applying sensing actions arbitrarily or using sensors that do not inform the agent about anything relevant for \( \phi. \)
}

\begin{example}\label{ex:missing knowledge and action} Let us assume quenching comes with an additional condition that it only works with metals: \(
	\Box Poss(quench(x)) \equiv Holding(x) \land Metal(x).
\) Let \( \inits \) be as before, and let the initial state of the world be given by \(
	\init\su * = \inits \cup \set{\neg Fragile(h), \neg Glass(h), Metal(h)}.
\) 
Consider \( \delta = pickup(h)\cdot drop(h) \), and it is easy to see that \(  \init\su *\land \dyn \land \oknow(\inits \land \dyn) \) entails \( [\delta] \neg \know Broken(h) \). Consider \( \delta' = pickup(h)\cdot quench(h)\cdot drop(h) \) which adds the quenching action. In itself, the sequence is not known to be executable because the agent does not know that \( h \) is metallic. So  \( \alpha = Metal(h) \) together with \( \delta' \) is the explanation because $\init\su *\land \dyn \land \oknow(\inits \land \alpha \land \dyn) $ entails 
 \(  [\delta'] (\phi\land \know\phi) \) where \( \phi = Broken(h) \), as well as the legality of the sequence and knowledge of its legality. 
	
\end{example}


\subsection{Agents Only-Knowing Weakened Truths} 
\label{sub:agents_only_knowing_weakened_truths}

Here we assume \( \inits \not\subseteq \init \) but for every \( \alpha\in\inits \), \( \init \models \alpha. \) In other words, 	\( \init \models \inits. \) 
That is, we might have an atom \( p\in\init, \) but \( \inits \) instead has \( (p\lor q). \) We then can define an account involving missing knowledge and actions, and so the same definition from \ref{defn missing knowledge and action} applies.

\begin{example} Let us consider  Example \ref{ex:missing knowledge and action} except that the initial theory of the agent is \( \init'' = \inits \land (Metal(h) \lor Metal(d)) \). That is, it includes all the formulas from \( \inits \) but also information that \( h \) is metallic or (falsely) that \( d \) is metallic. However, \( \init\su * \models (Metal(h) \lor Metal(d)) \), and so \( \alpha \) and \( \delta' \) from Example \ref{ex:missing knowledge and action} counts as the explanation. 
	
\end{example}




\subsection{Agents with False Beliefs} 
\label{sub:agents_with_false_beliefs}

False beliefs are only satisfiable when \( \Box (\know \alpha \supset \alpha) \) is not valid, as it happens to be in our case. 
For simplicity, we deal with the case of missing knowledge, and this can be easily coupled with missing actions in the manner discussed above. (Our example will deal with both.)

\begin{definition}\label{defn false beliefs}
	Suppose \( \inits \not\subseteq \init, \) and moreover \( \init\not\models \inits. \) Suppose \( \theory \not \models ([\delta]\know\phi \land \know Exec(\delta)) \) but \( \theory \models  Exec(\delta) \land [\delta]\phi. \) Suppose there is \( \alpha\in\init \) and \( \beta\in\inits \) such that \( \init\cup \dyn\cup \oknow( (\inits - \set{\beta}) \cup \set{\alpha} \cup \dyn ) \models [\delta]\know\phi \land \know Exec(\delta). \) Then the smallest such \( \alpha \) and \( \beta \)  are the explanations. 
\end{definition}

\begin{example} Consider Example \ref{ex:missing knowledge and action} and let \( \init'' = \init' \land \neg Metal(h) \). So the agent only-knows everything from  \( \init' \) as well as a false fact about \( h. \) Let \( \init\su *, \delta,  \)  and \( \delta' \) be  as in Example \ref{ex:missing knowledge and action}. Now note that given \( \alpha = Metal(h) \), \( \beta = \neg Metal(h) \) and \( \delta' \), we have \( \init\su * \land \dyn\land \oknow(\init'\land \alpha\land \dyn) \) entails \(
	 [\delta'](\phi \land \know\phi) \land Exec(\delta') \land \know Exec(\delta'), 
\)
for \( \phi = Broken(h) \). As it turns out \( \init' \) is obtained by removing \( \beta \) from \( \init'' \) by construction, and 
so \( \alpha,\beta  \) and \( \delta' \) constitutes as the explanation. 
	
\end{example}

\subsection{Possibility vs Knowledge} 
\label{sub:possibility_vs_knowledge}

In many applications, we may not require that the agent knows \( \phi \), only that it considers \( \phi \) possible. In the explanation generation framework of  \cite{vasileiou2022logic}, for example, there is a notion of credulous entailment where a single belief state suffices to checking the validity and achievement of a plan. (In contrast, skeptical entailment is when every belief state is involved.) The analogous notion in an epistemic language is to introduce a companion modal operator \( \Bel \) with the following semantics: \begin{itemize}
	\item \( e,w,z\models \Bel\alpha \) iff there is some \( w' \sim\sub z w, w'\in e \) such that \( e,w',z\models \alpha. \)
\end{itemize}
As long as there is at least one world where \( \alpha \) is true, \( \Bel\alpha \) is evaluated to true at \( (e,w,z) \). This is a companion  modal operator to \( \know \) for \emph{possibility.}  We might introduce an analogue to Definition \ref{defn missing knowledge and action} in using \( \Bel \) instead of \( \know \) as the modality in the goal. To see how this works, let us revisit Example \ref{ex:missing knowledge and action}. 

\begin{example} Consider the modified precondition axiom for \( quench(x) \), and let \( \inits \), \( \init\su * \), \( \delta \) and \( \delta' \) be as in Example \ref{ex:missing knowledge and action}. By only-knowing \( \inits \), the agent considers some worlds where \( h \) is metallic, and others where it is not. Thus, we now see that we do not really need to suggest \( \alpha = Metal(h) \) to the agent if the weaker notion of a CF explanation is considered. Indeed, \( \init\su * \land \dyn \land \oknow(\inits\land \dyn) \) entails \( [\delta'](\phi\land \Bel\phi) \) for \( \phi = Broken(h) \).

\end{example}

In other words, we have augmented actions but we did not need to augment knowledge in the above example.  Had the agent believed false things, then we might have needed to augment both, and so can appeal to Definition \ref{defn false beliefs} but using \( \Bel \) instead of \( \know \) to allow for credulous-type reasoning. 


\section{Other related efforts} 
\label{sec:other_related_efforts}

In addition to the works discussed in previous sections, the following efforts are  related.

There is some syntactic (and perhaps intuitive) connection to explanation-based diagnosis. For example, in \cite{sohrabi2010diagnosis}, the idea is to encode the behavior of the system to be diagnosed as a situation calculus action theory, encode observations as situation calculus formulae, and conjecture a sequence of actions to explain what went wrong with the system. (However, they often need to model  faulty or abnormal components when defining the notion of a diagnosis.)
In our setting, in contrast, we identify actions and/or knowledge that determine how an outcome can be changed. Nonetheless, we believe that a further formal study to relate such accounts would be useful, and could nicely complement empirical works such as \cite{dai2022counterfactual}. See also Ginsberg \cite{ginsberg1986counterfactuals}.

We previously discussed the reduction of projection and reasoning about knowledge to non-modal reasoning \cite{LakemeyerLevesque2004}, but we did not elaborate on generating plans. For more information on synthesizing plans, programs, and epistemic plans, see \cite{classen2007towards,DBLP:journals/logcom/DitmarschHL11,reifsteck2019epistemic,baral_et_al:DR:2017:8285} and their references.

An alternative approach to computing properties and plans is through answer set programming (ASP) \cite{baral2005logic,gelfond1993representing}, which also supports  reasoning about knowledge   \cite{fandinno2022thirty}. In general, our formalization does not preclude consideration of other logical languages. For example, in the simplest setting in this paper, a counterfactual explanation is the synthesis of a course of action that negates the goal or knowledge about the goal. In fact, \cite{bogatarkan2020explanation} consider counterfactual explanations for multi-agent systems, that is also motivated in terms of offering an  alternative course of action. Although they do not explore a range of definitions with references to knowledge as we do, exploring whether our definitions can be implemented in such approaches is worthwhile.

Such a course for formalisation may help better relate our efforts to declarative approaches to counterfactual explanations. For example,  \cite{bertossi2021declarative} explores the use of ASP for generating counterfactual explanations, but in a classical machine-learning sense, determined by how much certain features affect the overall prediction (understood as a causal link).


\section{Conclusions} 
\label{sec:conclusions}

We developed an account of counterfactual explanations and reconciliation-based counterfactual explanations in this paper. This allows for a simple and clear specification  in the presence of missing  actions, partial knowledge, weakened beliefs and false beliefs. Existing accounts of discrepancy in plans,  among others, can be seen as variations of this more general specification.  For the future, it would be interesting to incorporate other notions  in our formalization, such as operational aspects of plans, costs, optimality and conciseness  \cite{fox2017explainable,sreedharan2018handling,vasileiou2022logic}, towards a unified mathematical specification of explainable planning.

\bibliography{iclp}

\end{document}